\documentclass[11pt]{article}

\usepackage[preprint]{acl}
\usepackage{times}
\usepackage{latexsym}
\usepackage[T1]{fontenc}
\usepackage[utf8]{inputenc}
\usepackage{microtype}
\usepackage{inconsolata}
\usepackage{graphicx}
\usepackage{amsmath}
\usepackage{amssymb}
\usepackage{booktabs}
\usepackage{dblfloatfix}
\usepackage{placeins}
\usepackage{pgfplots}
\usepackage{subcaption}
\usepgfplotslibrary{groupplots}
\pgfplotsset{compat=1.18}
\usepackage{float}
\usepackage{enumitem}
\usepackage{xcolor}
\usepackage{xspace}

\newcommand{\system}{DocNavRAG\xspace}
\usepackage{booktabs}
\usepackage{tabularx}
\usepackage{array}
\usepackage{booktabs}
\usepackage{tabularx}
\usepackage{array}
\usepackage{ragged2e}
\usepackage{enumitem}

\usepackage{booktabs}
\usepackage{multirow}
\usepackage{xcolor}
\usepackage{colortbl}
\usepackage{svg}

\usepackage{cuted}

\usepackage{amsmath}

\definecolor{backbonegain}{HTML}{B3475A}
\definecolor{backboneloss}{HTML}{555B61}
\colorlet{backbonewine}{backboneloss}

\usepackage[textsize=scriptsize]{todonotes}
\usepackage[T1]{fontenc}
\newcounter{algorithm}

\title{\system: Document-Structured Graph RAG with Stateful Evidence Construction for Complex Document Question Answering}

\author{
\textbf{Dongyang Xie}\textsuperscript{1}
\quad
\textbf{Yao Tian}\textsuperscript{2}
\quad
\textbf{Hao Zhang}\textsuperscript{3}
\quad
\textbf{Yifei Yuan}\textsuperscript{4,*}
\\
\textbf{Tieyun Qian}\textsuperscript{1}
\quad
\textbf{Ming Zhong}\textsuperscript{1}
\quad
\textbf{Jiawei Jiang}\textsuperscript{1}
\quad
\textbf{Yuanyuan Zhu}\textsuperscript{1,*}
\\[1.2ex]
\textsuperscript{1}School of Computer Science, Wuhan University
\\
\textsuperscript{2}The Hong Kong University of Science and Technology
\\
\textsuperscript{3}The Chinese University of Hong Kong
\\
\textsuperscript{4}ETH Zurich
\\[0.8ex]
\small
\texttt{
\{2021302111223,qty,clock,jiawei.jiang,yyzhu\}@whu.edu.cn
}
\\
\texttt{ytianbc@cse.ust.hk}
\quad
\texttt{zhanghaowuda12@gmail.com}
\quad
\texttt{yuanyif@ethz.ch}
}

\begin{document}

\maketitle

% 第一页底部的通讯作者说明。
\begingroup
\renewcommand{\thefootnote}{\fnsymbol{footnote}}
\footnotetext[1]{
Corresponding authors: Yifei Yuan and
Yuanyuan Zhu.
}
\endgroup

\begin{abstract}

Answering complex questions over large document collections requires assembling complementary evidence across sections and documents. Existing Retrieval-Augmented Generation (RAG) approaches combine structured retrieval with adaptive reasoning, but typically rely either on fixed graph traversal or on weakly structured interfaces. On this basis, we propose an agentic framework that learns to navigate document structure within and across documents rather than repeatedly search from scratch. We introduce \system, which organizes document hierarchies and cross-region relationships into a navigable graph and exposes graph operations for locating, navigating, expanding, and fetching within an agentic framework. Throughout retrieval, \system maintains an evolving evidence state that guides exploration until sufficient evidence has been gathered.
We show that across four long- and multi-document QA benchmarks, \system improves answer quality and context sufficiency over the strongest baseline by 7.8\% and 17.7\% on average.

\end{abstract}
\section{Introduction}

Complex document question answering (CDQA) requires answering questions over collections of long documents, such as scientific papers, technical manuals, and government reports~\citep{li2024m3sciqa,huang-etal-2025-towards-multi,bai2025longbenchv2}. The required evidence is often scattered across multiple sections and documents, making it impractical for modern large language models (LLMs) to process entire corpora, especially when real-world collections exceed million-token context windows.
 Retrieval-augmented generation (RAG) provides a scalable alternative by partitioning the corpus into retrievable units and selecting a small subset as evidence for answer generation~\citep{lewis2020rag,gao2024ragsurvey}. They rely on flat, one-shot retrieval, overlooking corpus structure and iterative refinement~\citep{xiang2025whentousegraphs,jiang-etal-2023-active}. As a result, retrieved passages are often individually relevant but collectively insufficient for complex questions~\citep{joren2025sufficient}. 
 % The central challenge in CDQA is therefore constructing a compact context with sufficient, complementary evidence.

\begin{figure}[t]
\centering
\includegraphics[width=\columnwidth]{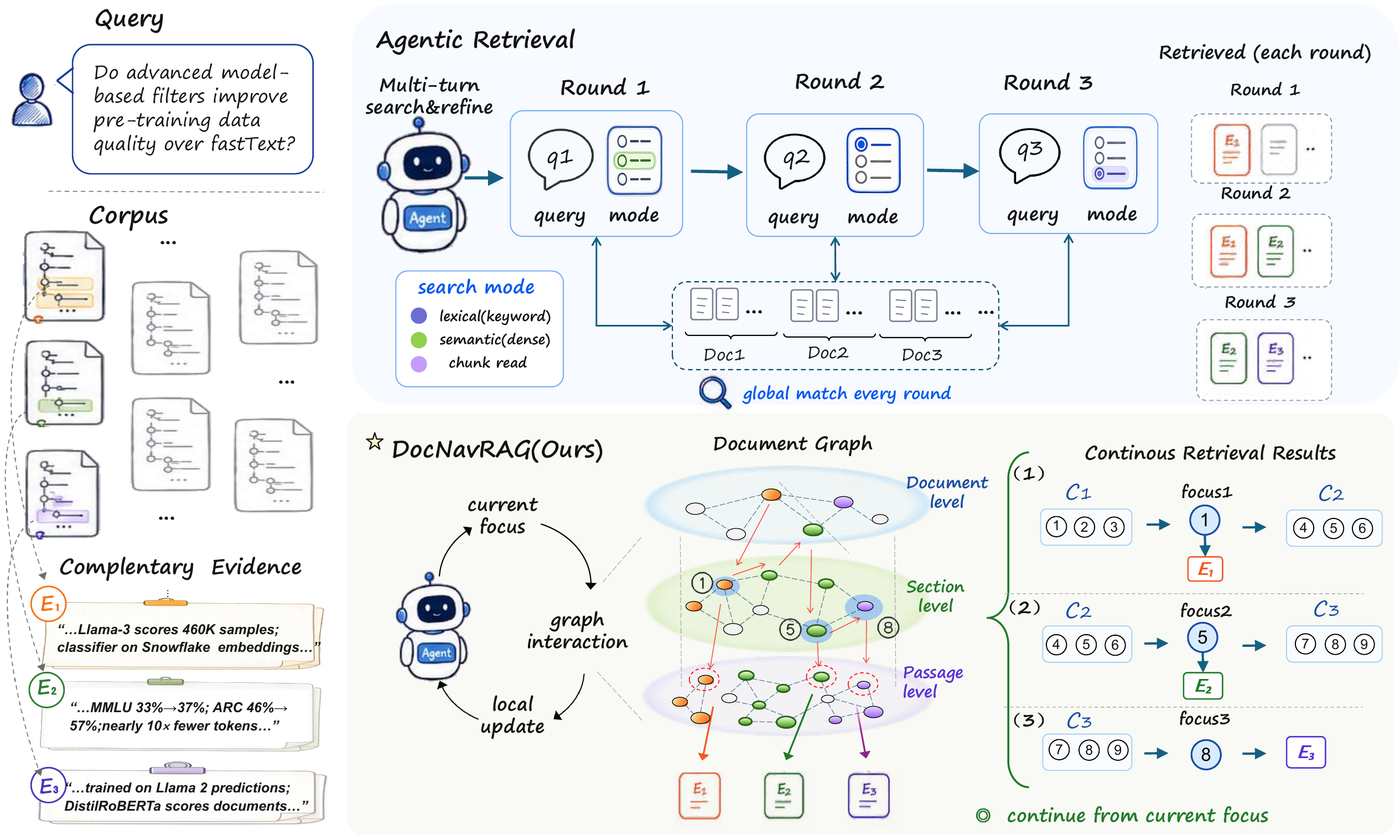}
\caption{
 Comparison of conventional agentic retrieval with \system's graph-based  retrieval in CDQA.
 }
\label{fig:intro-task-proposal}
\end{figure}

Existing methods address this challenge from two directions: improving corpus organization~\citep{edge2024graphrag,tao-etal-2025-treerag} and enhancing retrieval control~\citep{jiang-etal-2023-active,asai2024selfrag,du2026arag}. The former organizes document chunks and their semantic relationships into structured representations such as graphs, allowing retrieval to reach evidence beyond passages directly matched to the query. However, these methods typically rely on predefined retrieval or traversal strategies, limiting their ability to adapt evidence selection to the evolving information needs of a question. The latter leverages agentic RAG workflows that formulate retrieval actions and iteratively refine the collected evidence. 
% ; RL-based methods such as Search-o1 and Graph-R1 further learn the retrieval policy itself~\citep{li2025searcho1, luo2025graphr1}. 
However, the underlying corpus interfaces remain structurally limited: flat passage collections provide few navigation cues, while LLM-extracted knowledge graphs can be brittle and often lose the document-level hierarchy needed for high-level routing. Consequently, agents repeatedly issue global queries rather than move from broad document regions to precise evidence.

Motivated by this, as shown in Figure \ref{fig:intro-task-proposal}, we propose to expose source-document structure as a navigable search space, allowing the agent to move from partial evidence to what is still missing. We introduce \system, a training-free framework comprising a lightweight graph-based corpus organization layer and an agentic retrieval controller that navigates this graph to construct evidence adaptively. At indexing time, \system reconstructs the native structure of source documents as a multi-granular hierarchy while retaining original paragraphs as reliable evidence units. It then adds lightweight horizontal links between related or adjacent topics, both within and across documents. The resulting graph supports hierarchical navigation and associative expansion without requiring costly semantic graph extraction. At query time, a retrieval agent interacts with the graph through a set of graph operations, dynamically adjusting the search direction and granularity based on the query and accumulated evidence. Each retrieved passage updates the evidence state and guides subsequent actions until a sufficient evidence set is assembled.

We evaluate \system on four CDQA benchmarks containing multi-document and long-form scientific, technical, and government texts. \system outperforms passage-based, graph-based, and agentic RAG baselines on all benchmarks, improving answer quality over the strongest baseline by\% points on average. Further analyses show that \system remains robust across diverse corpus settings and LLM backbones while combining the strongest answer quality with an intermediate LLM inference cost. Our contributions are  as follows:
\begin{itemize}[leftmargin=12pt]
\item We introduce the first training-free agentic Graph-based RAG system for CDQA. The model adaptively orchestrates graph operations to retrieve focused, sufficient evidence for answer generation.

\item To facilitate agentic planning, we construct a lightweight document graph that captures document-native structure, multiple granularities, and cross-region relations, enabling structured navigation within and across documents.

\item Experiments across four benchmarks and six LLM backbones show consistent gains in answer quality and retrieved-context sufficiency compared with baseline methods, with average improvements of 7.8\% and 17.7\%, respectively.
\end{itemize}

\section{Related Work}

\paragraph{Graph-based RAG.}
Compared with standard RAG~\citep{lewis2020rag,gao2023ragsurvey}, Graph-based RAG  further indexes the source corpus as an interconnected graph and performs graph traversal to utilize structural reasoning~\citep{peng2024graphragsurvey}. One line of recent work enhances graph representations through multi-granular structures~\citep{edge2024graphrag,huang-etal-2025-retrieval,xu2025noderag} and higher-order relationships~\citep{luo2025hypergraphrag,wang2025hgrag,feng2025hyperrag}. Others improves graph traversal through propagation-based methods~\citep{gutierrez2024hipporag,zhuang2025linearrag}, path-oriented strategies~\citep{mavromatis-karypis-2025-gnn,chen2025pathrag}, or subgraph refinement~\citep{hu2024grag,wu2026toporag}. 
% These studies collectively demonstrate the value of leveraging graphic corpus structure to support complex reasoning.

\paragraph{Agentic RAG.}

Unlike standard RAG system that typically follows a predefined retrieval workflow~\citep{lewis2020rag,gao2023ragsurvey}, Agentic RAG incorporates retrieval control into the model's inference process~\citep{liang-etal-2025-reasoning}, allowing the retrieval workflow to evolve with the information gathered during execution. Modern agentic systems often combine planning and iterative reflection: planning organizes the information needs of a question into retrieval and reasoning subgoals~\citep{lee-etal-2024-planrag,verma2024planstarrag}, while intermediate results are used to assess the current evidence, revise the plan, and determine the next search direction~\citep{trivedi-etal-2023-interleaving,jiang-etal-2023-active,asai2024selfrag}. 
Recent systems have further expanded the range of retrieval tools available to the model, enabling it to not only decide what to retrieve but also select or combine operations~\citep{du2026arag,shen-etal-2025-gear}. We propose integrating a hierarchical document graph into an agentic retrieval framework.
% Together, these advances highlight the value of agentic control and suggest that broader, more flexible retrieval interfaces can further expand the capabilities of agentic systems.

\section{Problem Formulation}
\label{sec:preliminaries}

\begin{figure*}[t]
    \centering
    \includegraphics[width=\textwidth]{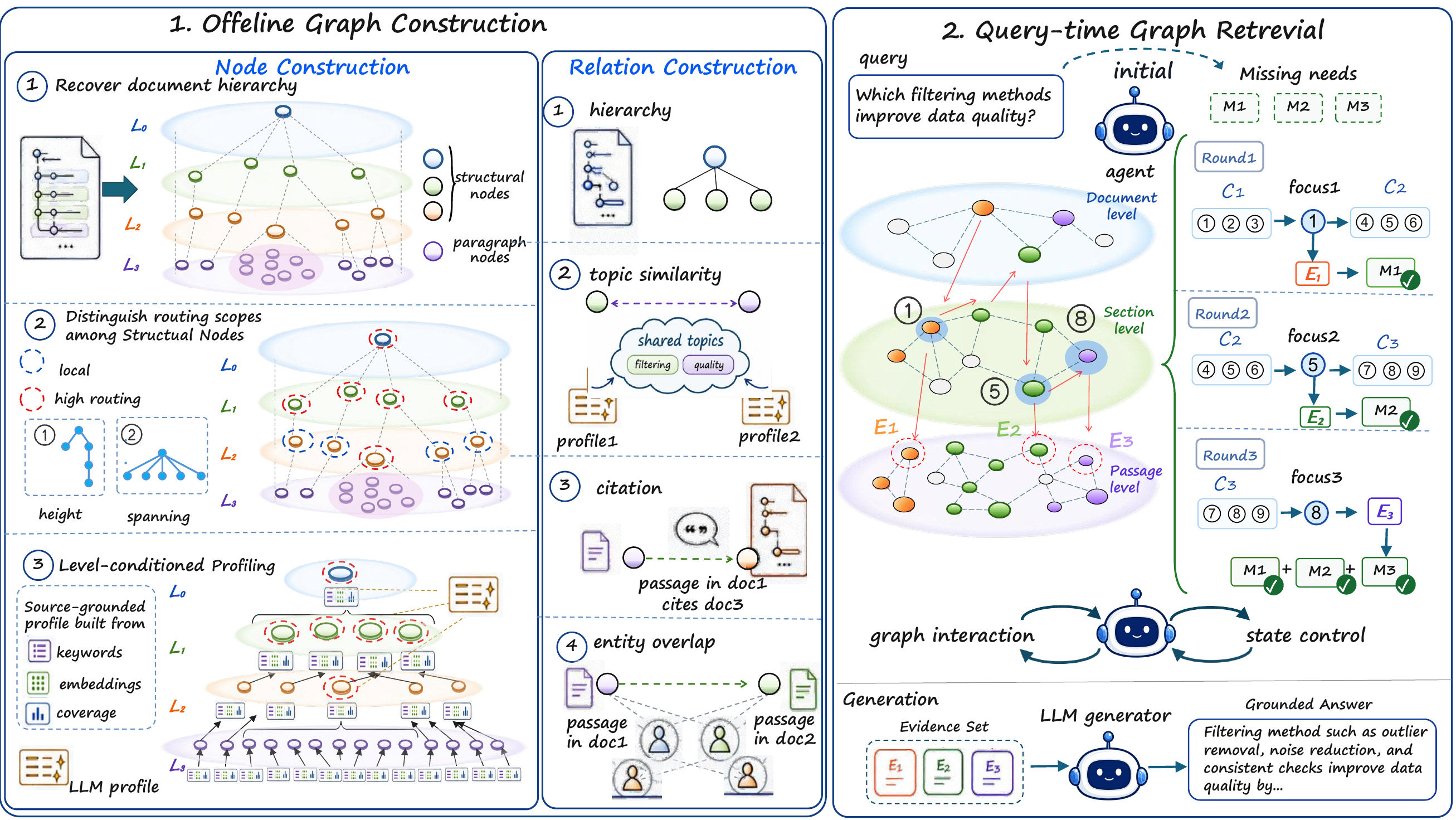}
    \caption{Framework overview of \system. Source documents are first organized into a multi-granular document graph by recovering document-native hierarchies, assigning level-conditioned node views, and adding structural and lateral relations. At query time, a retrieval agent iteratively locates, navigates, expands, and fetches graph nodes, updating its evidence state until sufficient source evidence is collected for answer generation.}
    \label{fig:framework-overview}
\end{figure*}

In CDQA, the input consists of a
question \(q\) and a collection of source documents
\(D=\{d_i\}_{i=1}^{n}\). The objective is to generate an answer \(y\)
grounded in the information contained in the document collection.
We decompose this process into three stages:
(1) corpus organization, (2) evidence construction, and (3) answer
generation. Formally, the overall pipeline can be expressed as:
\begin{equation}
y
=
\operatorname{Generate}
\!\left(
q,\,
\operatorname{Select}
\!\left(
q,\,
\Phi(D)
\right)
\right),
\label{eq:overall}
\end{equation}
where \(\Phi(\cdot)\) organizes the document collection into a
retrievable corpus interface, \(\operatorname{Select}(\cdot)\)
constructs a compact evidence set for the question, and
\(\operatorname{Generate}(\cdot)\) produces the final answer.

Each document is represented as an ordered sequence of source text
units,
$d_i=(u_{i,1},\ldots,u_{i,m_i})$,
where \(u_{i,j}\) denotes the \(j\)-th source unit of document
\(d_i\). Let
$\mathcal{U}
=
\{u_{i,j}\mid 1\le i\le n,\;1\le j\le m_i\}$
denote the set of all source units contained in the document
collection \(D\).

\paragraph{Corpus Organization.}

The document collection is first transformed into a retrieval interface,
\begin{equation}
\mathcal{I}
=
\Phi(D),
\label{eq:corpus-organization}
\end{equation}
where \(\mathcal{I}\) denotes a retrieval substrate over
\(\mathcal{U}\). This abstraction covers flat passage indexes,
hierarchical indexes, graph-structured corpora, and other corpus
organizations without assuming a specific retrieval structure.

\paragraph{Evidence Construction.}

Given the question \(q\), the system constructs a compact evidence set
from the retrieval interface,
\begin{equation}
E
=
\operatorname{Select}(q,\mathcal{I}),
\qquad
E\subseteq\mathcal{U},
\label{eq:evidence-construction}
\end{equation}
where \(E\) contains the source units required to answer the question
while excluding irrelevant or redundant information.

\paragraph{Answer Generation.}

Finally, the answer is generated conditioned on the selected evidence using an LLM,
\begin{equation}
y
=
\operatorname{Generate}(q,E).
\label{eq:answer-generation}
\end{equation}

\section{Method}

\subsection{Framework Overview}

As demonstrated in Figure~\ref{fig:framework-overview}, \system consists of two stages: offline document graph construction and query-time agentic retrieval for answer generation. Specifically, during indexing, the source corpus is transformed into a multi-granular document graph that preserves document-native organization while exposing additional relations between content regions. During querying, a retrieval agent interacts with the graph through a constrained action interface, updates its retrieval state over multiple rounds, and accumulates fetched source content as evidence for answer generation.

\subsection{Document Graph Construction}

We construct the document graph in three stages: recovering the hierarchy of each source document, constructing level-conditioned node profiles, and adding structural and lateral relations.

\subsubsection{Document Hierarchy Recovery}

Given a document corpus $D=\{d_i\}_{i=1}^{n}$, we recover a rooted, ordered hierarchy tree $T_d=(V_d,\mathcal{E}_d^{\mathrm{hier}})$ for each document $d\in D$
using its layout, headings, and reading order. Its node set is partitioned as
\begin{equation}
V_d=P_d\mathbin{{\cup}}S_d,
\label{eq:hierarchy-nodes}
\end{equation}
where $P_d$ and $S_d$ denote the paragraph and structural nodes of document $d$, and $\mathcal{E}_d^{\mathrm{hier}}$ contains the parent--child edges of the recovered hierarchy.

Each paragraph node corresponds to a source paragraph and forms a leaf, whereas structural nodes represent nested document scopes such as sections, chapters  and the entire document. These nodes are organized by scope containment, with paragraph nodes attached to their narrowest enclosing scope and siblings retaining the original reading order, providing the structural basis for the node profiling and graph relations introduced below.

\subsubsection{Level-conditioned Node Profiling}

Structural nodes at different levels cover source content of varying scope and granularity. Locating content within a narrow scope typically requires fine-grained local signals, whereas choosing among broader search regions relies more on representations that summarize their overall content. We therefore distinguish structural scopes used for higher-level routing from more local scopes and construct their retrieval profiles accordingly.

\paragraph{Routing-Scope Selection.}
For a structural node $u\in S_d$, let $h(u)$ denote the height of its subtree and $\ell(u)$ denote the total length of the source text covered by the node. Given thresholds $\tau_h$ and $\tau_\ell$, $R_d\subseteq S_d$ denotes the set of structural nodes selected as high-level routing scopes for document $d$, 
\begin{equation}
\begin{aligned}
R_d
&=
\{\operatorname{root}(T_d)\} \\
&\quad\cup
\{u\in S_d
\mid h(u)\geq\tau_h
\lor \ell(u)\geq\tau_\ell\}.
\end{aligned}
\label{eq:routing-scopes}
\end{equation}
Nodes in $R_d$ represent broader document regions and provide routing signals for higher-level navigation, while the remaining structural nodes retain profiles within more local scopes. This selection changes only the profile associated with each node and leaves the nodes and hierarchical relations of the document tree unchanged.

\paragraph{Profile Construction.}
To support retrieval at different document granularities, each node requires a representation of the source content within its scope. 
For each document $d\in D$, we first construct a source-grounded profile $\psi(v)$ for every node $v\in V_d$. For a paragraph node $v\in P_d$, $\psi(v)$ retains its original text, embedding, and structural position. For a structural node $v\in S_d$, $\psi(v)$ is constructed bottom-up by aggregating the corresponding signals.  
We then obtain the final retrieval profile  $\phi(v)$ by augmenting the nodes in $R_d$ with an LLM-generated summary: 
\begin{equation}
\phi(v)=
\begin{cases}
\psi(v),
& v\in V_d\setminus R_d, \\[2pt]
\bigl(\psi(v),m_v\bigr),
& v\in R_d,
\end{cases}
\label{eq:node-profile}
\end{equation}
where $m_v$ summarizes the source content covered by node $v$. Thus, all nodes retain profiles derived from the underlying source text, while semantic summarization is introduced only at structural scopes used for higher-level routing. Detailed profile components and bottom-up aggregation procedures are provided in Appendix~\ref{app:graph-construction-details}.

\subsubsection{Graph Relation Construction}

The recovered document trees preserve the corpus's native structure. We include two complementary graph relations: \emph{structural edges}, which preserve this hierarchy, and \emph{lateral edges}, which connect related regions across it.

\paragraph{Structural Edges.}
The parent-child relations in each document tree form a set of hierarchical edges $\mathcal{E}_{\mathrm{hier}}$, enabling retrieval to move between paragraph nodes and their enclosing structural scopes. We additionally construct local edges $\mathcal{E}_{\mathrm{local}}$ from reading order and structural adjacency, connecting consecutive regions within the same parent scope. Together, these document-native relations form
\begin{equation}
\mathcal{E}_{\mathrm{struct}}
=
\mathcal{E}_{\mathrm{hier}}
\cup
\mathcal{E}_{\mathrm{local}}.
\label{eq:structural-relations}
\end{equation}

\paragraph{Lateral Edges.}
Beyond document-native structure, we construct four main types of lateral edges from citation, entity, semantic, and community signals to provide direct access to related regions globally, defined as
\begin{equation}
\begin{aligned}
\mathcal{E}_{\mathrm{lat}}
&=
\mathcal{E}_{\mathrm{cite}}
\cup
\mathcal{E}_{\mathrm{ent}}\cup
\mathcal{E}_{\mathrm{sim}}
\cup
\mathcal{E}_{\mathrm{comm}}.
\end{aligned}
\label{eq:lateral-relations}
\end{equation}
Table~\ref{tab:lateral-relations} summarizes the signal and endpoint rule of each relation type. Detailed edge-construction and endpoint-selection procedures are provided in Appendix~\ref{app:graph-construction-details}.

\begin{table}[t]
\centering
\footnotesize
\setlength{\tabcolsep}{3pt}
\renewcommand{\arraystretch}{1.16}

\begingroup
\hyphenpenalty=10000
\exhyphenpenalty=10000
\renewcommand{\tabularxcolumn}[1]{m{#1}}

\begin{tabularx}{\columnwidth}{
    @{}
    >{\centering\arraybackslash}m{0.14\columnwidth}
    >{\RaggedRight\arraybackslash}m{0.22\columnwidth}
    >{\RaggedRight\arraybackslash}X
    @{}
}
\toprule
\textbf{Relation}
& \textbf{Signal}
& \textbf{Connection Rule} \\
\midrule

$\mathcal{E}_{\mathrm{cite}}$
& Citation
& Citing paragraph and its structural ancestors
  $\rightarrow$ cited document node \\

\addlinespace[1.5pt]

$\mathcal{E}_{\mathrm{ent}}$
& Entity overlap
& Matched-granularity nodes sharing salient entities \\

\addlinespace[1.5pt]

$\mathcal{E}_{\mathrm{sim}}$
& Semantic similarity
& Matched-granularity nodes with high semantic similarity \\

\addlinespace[1.5pt]

$\mathcal{E}_{\mathrm{comm}}$
& Community membership
& High-level structural nodes within the same topical community \\

\bottomrule
\end{tabularx}
\endgroup

\caption{Lateral relation signals and connection rules.}
\label{tab:lateral-relations}
\end{table}

Combining the node sets introduced above with the structural and lateral
relations, we obtain the final document graph:
\begin{equation}
\begin{aligned}
V
&=
P\cup S, \\
\mathcal{E}
&=
\mathcal{E}_{\mathrm{struct}}
\cup
\mathcal{E}_{\mathrm{lat}}, \\
G
&=
(V,\mathcal{E}).
\end{aligned}
\label{eq:document-graph}
\end{equation}
We next describe how the retrieval agent interacts with $G$ to gather source evidence.

\FloatBarrier
\subsection{Agentic Retrieval over Graph Substrate}

Given a question $q$ and a document graph $G$, the retrieval agent incrementally gathers evidence through a constrained graph-action interface that accesses local graph regions according to the current retrieval state. We first define the available operations and retrieval state, followed by the multi-round retrieval process.

\subsubsection{Retrieval Agent Setup}

\paragraph{Graph Access Operations.}
We define an action space $\mathcal{A}$ comprising four complementary graph access operations. Table~\ref{tab:graph-actions} summarizes the graph component accessed and the nodes returned by each operation. Implementation details are provided in Appendix~\ref{app:agentic-workflow}.

\begin{table}[t]
\centering
\footnotesize
\setlength{\tabcolsep}{3.5pt}
\renewcommand{\arraystretch}{1.14}

\begin{tabularx}{\columnwidth}{
    @{}
    >{\RaggedRight\arraybackslash}p{0.18\columnwidth}
    >{\RaggedRight\arraybackslash}p{0.27\columnwidth}
    >{\RaggedRight\arraybackslash}X
    @{}
}
\toprule
\textbf{Action}
& \textbf{Accesses}
& \textbf{Returns} \\
\midrule

\textsc{Locate}
& Node profiles
& Relevant nodes at selected granularities \\

\textsc{Navigate}
& Hierarchical edges
& Nodes in broader or narrower document scopes \\

\textsc{Expand}
& Lateral edges
& Related nodes within or across documents \\

\textsc{Fetch}
& Node--source associations
& Associated source paragraph nodes \\

\bottomrule
\end{tabularx}

\caption{Graph access operations available to the retrieval agent.}
\label{tab:graph-actions}
\end{table}

\paragraph{Retrieval State.}
To retain information acquired in previous rounds, the agent maintains a compact retrieval state at round $t$:
\begin{equation}
s_t=(E_t,C_t,M_t,F_t).
\label{eq:retrieval-state}
\end{equation}

\begin{itemize}[
    leftmargin=1.25em,
    labelsep=0.45em,
    itemsep=0pt,
    parsep=0pt,
    topsep=2pt,
    partopsep=0pt
]
\item $E_t$: the paragraph nodes retained as source evidence.
\item $C_t$: the candidate graph nodes retained for later exploration.
\item $M_t$: the remaining evidence needs.
\item $F_t$: the node selected for the next local graph interaction;
$F_t=\varnothing$ denotes graph-wide search.
\end{itemize}
Together, these components record the current retrieval progress and provide the context for the next graph interaction.

\subsubsection{Adaptive Graph Interaction}

The agent infers the initial evidence needs $M_0$ from the
question and initializes the retrieval state as
\begin{equation}
s_0=(\varnothing,\varnothing,M_0,\varnothing).
\label{eq:initial-state}
\end{equation}
Starting from $s_0$, the agent repeatedly selects a graph action, processes its returned result to revise the retrieval state, and assesses whether the accumulated evidence is sufficient.

\paragraph{Step 1: Action Invocation.}

At round $t\geq1$, the retrieval agent $\mathcal{R}$ selects an action from $\mathcal{A}$ based on the question and the preceding retrieval state:
\begin{equation}
a_t=\mathcal{R}(q,s_{t-1};\mathcal{A}).
\label{eq:action-selection}
\end{equation}
The current focus determines the operating scope: $F_{t-1}=\varnothing$ denotes graph-wide interaction, whereas $F_{t-1}\in C_{t-1}$ identifies a selected node for local graph interaction. The action is then executed within this scope, producing an output $o_t$ for subsequent state revision.

\paragraph{Step 2: State Update over Intermediate Output.}

The selected action returns output $o_t$ of an initial set of paragraph or structural nodes. The agent then updates the retrieval state from the preceding state $s_{t-1}$ and the current action output $o_t$ through three updates. \noindent\textbf{(1) Result Screening.} The agent screens the nodes returned in $o_t$ according to the question and the preceding state. Paragraph nodes that directly address the
current evidence needs are retained as new evidence, while clearly
irrelevant paragraph or structural nodes are discarded. The remaining
nodes that may support further exploration are retained as new
candidates. We denote the retained evidence and candidate nodes by
$\Delta E_t$ and $\Delta C_t$, respectively, and merge them with the corresponding components from the preceding state. \noindent\textbf{(2) Need Revision.} The agent reassesses the evidence needs against the question and the
updated evidence. Satisfied needs are removed, while those that remain
unmet are retained or refined to form $M_t$. \noindent\textbf{(3) Focus Selection.} Based on $M_t$, the agent selects a single node from $C_t$ as the focus $F_t$ for the next local graph interaction. It sets
$F_t=\varnothing$ when the remaining needs call for graph-wide search. The resulting updates are summarized as
\begin{subequations}
\label{eq:state-update}
\begin{gather}
\begin{aligned}
(\Delta E_t,\Delta C_t)
&=
\operatorname{Screen}(o_t\mid q,s_{t-1}),\\
E_t
&=
E_{t-1}\cup\Delta E_t,\\
C_t
&=
C_{t-1}\cup\Delta C_t
\end{aligned}
\label{eq:result-screening}
\\
M_t
=
\operatorname{UpdateNeed}(M_{t-1}\mid q,E_t)
\label{eq:need-revision}
\\
F_t
=
\operatorname{SelectFocus}(C_t\mid M_t)
\label{eq:focus-selection}
\end{gather}
\end{subequations}

\paragraph{Step 3: Completion Assessment.}
After each state update, the agent reassesses the accumulated evidence and any unresolved information needs. Retrieval terminates when the step budget $B$ is exhausted or the collected evidence is deemed sufficient, i.e., $M_t=\varnothing$. Let $T$ denote the terminal step. The final answer is then generated from the question and the accumulated source evidence:
\begin{equation}
{y}=\operatorname{Generate}(q,E_T).
\label{eq:answer-generation}
\end{equation}
% where $\operatorname{Generate}$ denotes the LLM-based answer generator.

\section{Experiments}

\subsection{Experimental Setup}

\paragraph{Datasets.}
We evaluate on four CDQA benchmarks.
\textbf{(1) Large Document Collections.}
ScholarQA~\citep{asai2024openscholar} and MDAQA~\citep{huang-etal-2025-towards-multi} evaluate open-ended multi-document QA with collection-wide retrieval over 137 and 792 full-text scientific papers. \textbf{(2) Long Documents.} LongBench-v2 focuses on long-document QA~\citep{bai2025longbenchv2}. LongBenchV2-UserGuide evaluates option-level binary verification over 22 technical manuals, with each question grounded in a single document; LongBenchV2-Gov evaluates multiple-choice QA across 44 heterogeneous government documents, both of which contain fewer but substantially longer documents. Complete benchmark statistics are provided in Appendix Table~\ref{tab:dataset_statistics}.

 \paragraph{Baselines.} 
We compare against baselines across three RAG paradigms: 
\textbf{(1) Basic RAG.} BM25~\citep{robertson2009probabilistic}, TF-IDF~\citep{salton1988term}, and dense vector retrieval~\citep{reimers-gurevych-2019-sentence} instantiated with \texttt{sentence-transformers/all-mpnet-base-v2} cover sparse lexical and dense semantic matching. \textbf{(2) Graph-based RAG.}  LinearRAG~\citep{zhuang2025linearrag}, HippoRAG~\citep{gutierrez2024hipporag}, and LightRAG~\citep{guo2024lightrag} represent relation-free graph retrieval, knowledge-graph propagation, and hybrid graph--vector retrieval, respectively. \textbf{(3) Agentic RAG.} IRCoT~\citep{trivedi-etal-2023-interleaving} interleaves retrieval with reasoning, whereas A-RAG~\citep{du2026arag} provides an LLM controller with hierarchical search and reading interfaces.

\paragraph{Evaluation Metrics.} 
We evaluate both retrieved-context sufficiency and answer quality with the following metrics. 
\textbf{(1) Retrieved-Context Sufficiency.} 
Across all benchmarks, Sufficient Context assesses whether the retrieved evidence contains enough information to answer the question \citep{joren2025sufficient}, providing a common retrieval-side measure when consistent gold evidence annotations are unavailable.
\textbf{(2) Answer Quality.} 
For benchmarks with open-ended long-form responses, including ScholarQA and MDAQA, we report Answer Correctness~\citep{es2023ragas}, a metric that evaluates the semantic correctness of generated answers. We additionally report RAGAS Semantic Similarity~\citep{es2023ragas} as a complementary reference-based metric that measures the semantic overlap between the generated and reference answers. For LongBenchV2-UserGuide, we report option-level binary accuracy, while for LongBenchV2-Gov, we report multiple-choice accuracy. 
We additionally report RAGAS Faithfulness~\citep{es2023ragas}, Token-F1, and BERTScore-F1~\citep{zhang2020bertscore} for the open-ended benchmarks and list the results in Appendix Table~\ref{tab:appendix_secondary_diagnostics}.

% Fig. 1 and Fig. 2 are manual architecture/workflow figures.
% They are intentionally not generated by /paper-figure because they are not data-driven.

% === Table 1: Main Results ===
% =========================================================
% Colors
% =========================================================
\definecolor{MRScienceHead}{HTML}{EDF3F7}
\definecolor{MRLongHead}{HTML}{F5F0E6}
\definecolor{MRBestBase}{HTML}{DFF1EC}
\definecolor{MRBestBaseText}{HTML}{1E4F46}
\definecolor{MROursRow}{HTML}{EEE8F8}
\definecolor{MROursAccent}{HTML}{7F65AC}

% Strongest baseline:
% normal font weight + green background + underline
\newcommand{\mrbestbase}[1]{%
  \cellcolor{MRBestBase}%
  \textcolor{MRBestBaseText}{\underline{#1}}%
}

% Best overall result
\newcommand{\mrbestsystem}[1]{\textbf{#1}}

% =========================================================
% Main results table
% =========================================================
\begin{table*}[t]
\centering
\small
\setlength{\tabcolsep}{2.8pt}
\renewcommand{\arraystretch}{1.15}

\begin{tabularx}{
  \textwidth
}{
  @{}
  l
  *{10}{>{\raggedleft\arraybackslash}X}
  @{}
}

\toprule

\textbf{Method}
&
\multicolumn{6}{c}{
  \cellcolor{MRScienceHead}
  \textbf{Scientific QA}
}
&
\multicolumn{4}{c}{
  \cellcolor{MRLongHead}
  \textbf{LongBench-v2}
}
\\

\cmidrule(lr){2-7}
\cmidrule(lr){8-11}

&
\multicolumn{3}{c}{\textbf{ScholarQA}}
&
\multicolumn{3}{c}{\textbf{MDAQA}}
&
\multicolumn{2}{c}{\textbf{UserGuide}}
&
\multicolumn{2}{c}{\textbf{Gov.}}
\\

\cmidrule(lr){2-4}
\cmidrule(lr){5-7}
\cmidrule(lr){8-9}
\cmidrule(lr){10-11}

&
\textbf{Corr.}
&
\textbf{Sim.}
&
\textbf{Suff.}
&
\textbf{Corr.}
&
\textbf{Sim.}
&
\textbf{Suff.}
&
\textbf{Acc.}
&
\textbf{Suff.}
&
\textbf{Acc.}
&
\textbf{Suff.}
\\

% =========================================================
% Standard passage retrieval
% =========================================================
\midrule

\multicolumn{11}{l}{
  \textit{Standard passage retrieval}
}
\\[-1pt]

BM25
& 0.412
& 0.540
& 0.697
& 0.446
& 0.551
& 0.433
& 0.653
& 0.784
& 0.429
& 0.476
\\

TF-IDF
& 0.406
& 0.552
& 0.727
& 0.374
& 0.480
& 0.300
& 0.626
& 0.716
& 0.333
& 0.286
\\

Dense
& 0.389
& 0.514
& 0.758
& 0.416
& 0.511
& \mrbestbase{0.467}
& 0.589
& 0.614
& 0.286
& 0.286
\\

% =========================================================
% Graph-based RAG
% =========================================================
\midrule

\multicolumn{11}{l}{
  \textit{Graph-based RAG}
}
\\[-1pt]

LinearRAG
& 0.393
& 0.523
& 0.788
& 0.413
& 0.512
& 0.400
& 0.563
& 0.511
& 0.286
& 0.143
\\

HippoRAG
& 0.352
& 0.458
& 0.758
& 0.323
& 0.401
& 0.300
& 0.619
& 0.761
& 0.333
& 0.286
\\

LightRAG
& 0.413
& 0.543
& 0.454
& 0.336
& 0.409
& 0.400
& 0.479
& 0.489
& 0.286
& 0.429
\\

% =========================================================
% Agentic retrieval
% =========================================================
\midrule

\multicolumn{11}{l}{
  \textit{Agentic RAG}
}
\\[-1pt]

IRCoT
& 0.434
& 0.570
& 0.727
& 0.380
& 0.442
& 0.333
& 0.531
& 0.682
& 0.286
& 0.381
\\

A-RAG
& \mrbestbase{0.671}
& \mrbestbase{0.743}
& \mrbestbase{0.818}
& \mrbestbase{0.596}
& \mrbestbase{0.650}
& 0.367
& \mrbestbase{0.704}
& \mrbestbase{0.796}
& \mrbestbase{0.476}
& \mrbestbase{0.667}
\\

% =========================================================
% Proposed
% =========================================================
\midrule

% \multicolumn{11}{l}{
%   \textit{\textcolor{MROursAccent}{Proposed}}
% }
% \\
% [-1pt]

\rowcolor{MROursRow}
\textcolor{MROursAccent}{\textbf{\system}}
& \mrbestsystem{0.738}
& \mrbestsystem{0.797}
& \mrbestsystem{0.970}
& \mrbestsystem{0.618}
& \mrbestsystem{0.688}
& \mrbestsystem{0.800}
& \mrbestsystem{0.784}
& \mrbestsystem{0.886}
& \mrbestsystem{0.619}
& \mrbestsystem{0.800}
\\

\bottomrule

\end{tabularx}

\caption{
Main answer-quality and evidence results.
Corr.\ denotes RAGAS Answer Correctness for ScholarQA and MDAQA,
whereas Acc.\ denotes question-level accuracy for the two LongBench-v2
subsets.
Sim.\ denotes RAGAS Semantic Similarity, and Suff.\ denotes Sufficient
Context.
Bold values indicate the best overall result in each column;
green-shaded and underlined values indicate the strongest baseline.
}

\label{tab:main_results}
\end{table*}

\subsection{Main Results}

Table~\ref{tab:main_results} presents the end-to-end results. Overall, \system achieves the best performance across all reported answer-quality and retrieved-context sufficiency comparisons, with A-RAG notably emerging as the strongest baseline.
We draw two further observations from comparisons across retrieval paradigms:
\textbf{(1) Graph structure alone does not consistently translate into answer-quality gains.} Comparing the best-performing baseline in each group, graph-based RAG leads only on ScholarQA Correctness by 0.1 percentage points lightly, while passage retrieval leads in most datasets. 
\textbf{(2) Agentic retrieval benifits from the retrieval interface it operates over beyond query adaption.} Across all comparison over both answer-quality and retrieved-context sufficiency, performance consistently improves from IRCoT with iterative query updates, to A-RAG with hierarchical search operations, and further to \system with dynamic navigation over a document graph.

\subsection{Ablation Analysis}
We evaluate ablations of graph access and retrieval workflow with each benchmark's primary answer-quality metric. For \textbf{graph access}, \textit{w/o Hierarchy Search (HS)} restricts retrieval to local text units,
\textit{w/o Structural Navigation (SN)} removes navigation through hierarchical and local structural relations, and \textit{w/o Lateral Navigation (LN)} removes navigation over lateral links. For the \textbf{retrieval workflow}, \textit{w/o State Control (SC)} replaces evidence-conditioned control with a fixed retrieval sequence, while \textit{w/o Evidence Fetch (EF)} passes intermediate graph views directly to the generator without fetching source text.

\begin{figure*}[!t]
\centering
\includegraphics[width=0.9\textwidth]
{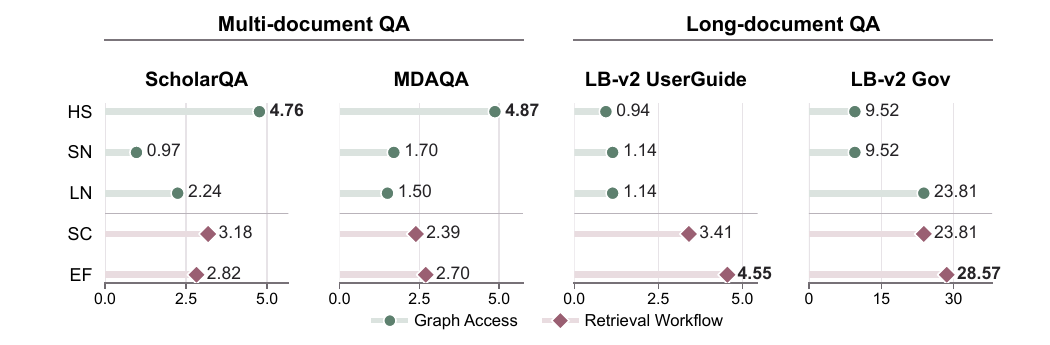}
\caption{
Per-benchmark ablation effects, measured as absolute drops (percentage points) in each benchmark's primary answer-quality metric relative to the full system.
Circles and diamonds denote graph-access and retrieval-workflow ablations, respectively; boldface identifies the largest drop.
Full results and details see Appendix~\ref{app:ablation-details}.
}
\label{fig:ablation_impact}
\end{figure*}

As shown in Figure~\ref{fig:ablation_impact}, removing any capability degrades performance, while the shared and setting-specific patterns across benchmarks show how \system draws on complementary capabilities to meet the retrieval demands of different corpus settings. In \textbf{Large document collections} such as  ScholarQA and MDAQA, retrieval spans large collections of moderately sized, relatively independent papers, making corpus-level localization the dominant challenge. Accordingly, \textit{Hierarchy Search} contributes most by expanding search coverage, whereas \textit{Structural Navigation} and \textit{Lateral Navigation} have smaller effects because these settings place less emphasis on deep within-document traversal or link-based navigation between already located regions.
In \textbf{long-document settings}, substantially longer individual documents in LongBench-v2 shift the dominant challenge from corpus-level localization to evidence completion across deep document regions. Accordingly, \textit{State Control} and \textit{Evidence Fetch} contribute most by sustaining evidence-guided search among topically similar local regions and recovering source details not retained in generalized higher-level representations, respectively. LongBenchV2-Gov further combines this depth with evidence distributed across heterogeneous sources, making \textit{Lateral Navigation} more important for reaching complementary evidence in other documents.

\subsection{Model Backbone Generalization}

We further evaluate \system with A-RAG across different LLM backbones to examine its robustness to backbone choice.
Table~\ref{tab:backbone_summary} summarizes this backbone-sensitivity analysis.

\begin{table}[t]
\centering
\footnotesize
\setlength{\tabcolsep}{3.2pt}
\renewcommand{\arraystretch}{1.07}

\begin{tabular*}{\columnwidth}{
    @{\extracolsep{\fill}}lcc@{}
}
\toprule
Backbone & Correctness & Similarity \\
\midrule
\textbf{Overall}
& \textbf{19/24} \; \textbf{(\textcolor{red}{+2.42})}
& \textbf{8/12} \; \textbf{\textcolor{red}{+1.20}} \\
\midrule
DeepSeek-v4-Flash$^\dagger$
& \textbf{4/4} \; \textbf{(\textcolor{red}{+6.60})}
& \textbf{2/2} \;  \textbf{(\textcolor{red}{+4.60})}\\
DeepSeek-v4-Pro
& \textbf{4/4} \; \textbf{(\textcolor{red}{+3.99})}
& 1/2 \; (+0.97) \\
\addlinespace[1pt]
Qwen3.5-Flash
& \textbf{4/4} \; \textbf{(\textcolor{red}{+4.11})}
& \textbf{2/2} \; \textbf{(\textcolor{red}{+0.86})} \\
Qwen3.5-Plus
& 2/4 \; (-1.14)
& 0/2 \; (-1.27) \\
\addlinespace[1pt]
GPT-5.4-mini
& \textbf{4/4} \; \textbf{(\textcolor{red}{+2.72})}
& \textbf{2/2} \; \textbf{(\textcolor{red}{+1.59})} \\
GPT-5.5
& 1/4 \; (-1.72)
& 1/2 \; \textbf{(\textcolor{red}{+0.44})} \\
\bottomrule
\end{tabular*}

\caption{
Backbone-wise comparison with A-RAG.
Each cell reports wins and the mean $\Delta$ in percentage points
in parentheses; $^\dagger$ marks the main-experiment backbone.
Full results see
Appendix Table~\ref{tab:backbone_generalization}.
}
\label{tab:backbone_summary}
\end{table}

Across 24 backbone-benchmark comparisons, \system outperforms A-RAG in 19 settings for Correctness, with an average gain of +2.42 points. Among the 12 Semantic Similarity comparisons reported on ScholarQA and MDAQA, it achieves higher scores in 8 settings, with an average gain of +1.20 points. The consistency of these gains across backbones demonstrates that the improvement is not specific to the main DeepSeek-v4-Flash configuration. Notably, the generally larger gains under the lighter configurations such as DeepSeek, Qwen3.5-Flash and gpt 5.4-mini configuration, highlighting that the document-side graph and constrained retrieval process remain useful even with less capable agent backbones.

\subsection{Cost Profiles across Retrieval Paradigms}
\begin{figure}[t]
\centering
\includegraphics[width=\columnwidth]{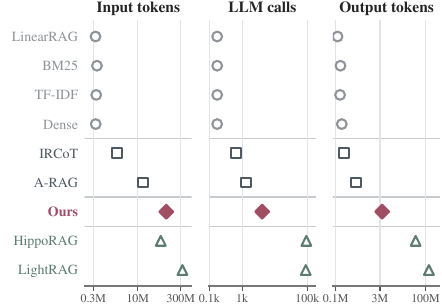}
\caption{
Aggregate end-to-end LLM usage across the four benchmarks. Details see Appendix Table~\ref{tab:cost_results}.
}
\label{fig:cost_profile}
\end{figure}

%Figure~\ref{fig:cost_profile} places \system in the middle of the inference-cost range. Viewed together with the main results, \system delivers higher answer quality at lower cost than semantic graph pipelines, while its additional inference over LLM-free indexing is accompanied by clear quality gains. These results highlight our approach as a quality-oriented design that achieves leading performance without incurring the highest inference cost.
Figure~\ref{fig:cost_profile} reveals a cost spectrum shaped by the extent of LLM involvement in corpus construction and query-time retrieval: LLM-free indexing occupies the lower end, LLM-intensive semantic graph construction the upper end, and \system lies between them through selective graph profiling and agentic navigation. Viewed together with the main results, this intermediate profile characterizes \system as a quality-oriented design whose additional inference supports clear quality gains without entering the highest-cost regime.

\section{Conclusion}

We presented \system, a training-free agentic GraphRAG framework for constructing focused and sufficient evidence in complex-document QA. It exposes source-document organization as a navigable search space and uses an evolving evidence state to guide retrieval from broad scopes to complementary source passages. Across four multi- and long-document QA benchmarks, \system consistently improves answer quality and retrieved-context sufficiency over passage-, graph-, and agentic retrieval baselines. Ablation, backbone, and cost analyses further show complementary roles for graph access and retrieval control across corpus settings, effectiveness across diverse LLM configurations, and an intermediate inference-cost profile.
\section*{Limitations}

This work focuses on source-document collections with recoverable organization, such as scientific papers, technical manuals, and government documents. \system builds on document-native signals, including layout, headings, reading order, citations, and topical relations, leaving more dynamic or strongly multimodal sources for future study. The graph is intended as a lightweight retrieval substrate rather than a fully induced semantic knowledge graph, preserving source structure and a clear boundary between navigation signals and fetched evidence. Agentic graph retrieval also adds query-time computation, although our cost analysis places \system between lightweight retrieval baselines and more expensive semantic graph pipelines; future work can reduce this overhead through caching, adaptive budgets, and smaller controller models.

\bibliography{references}
% \clearpage
\raggedbottom
\appendix
\section{Experiment Details}

\subsection{Datasets and Evaluation Units}
\label{app:datasets-evaluation-units}

Tables~\ref{tab:appendix_benchmarks} and~\ref{tab:dataset_statistics} provide dataset-level details that complement the experimental setup.
Table~\ref{tab:appendix_benchmarks} specifies the evaluation unit, source type, task setting, answer format, and primary metric for each benchmark.
Table~\ref{tab:dataset_statistics} reports the corresponding corpus scale.

\begin{table*}[!t]
\centering
\scriptsize
\setlength{\tabcolsep}{2.5pt}
\begin{tabular}{p{0.15\textwidth}p{0.14\textwidth}p{0.17\textwidth}cp{0.07\textwidth}p{0.17\textwidth}p{0.16\textwidth}}
\toprule
Dataset & Source type & Task setting & \# Qs. & \# Docs. & Answer format & Main correctness metric \\
\midrule
ScholarQA & CS/NLP scientific papers & Scientific multi-document QA & 33 & 137 & Open-ended long-form answer & RAGAS Answer Correctness \\
MDAQA & Scientific papers & Scientific multi-document QA & 30 & 792 & Open-ended long-form answer & RAGAS Answer Correctness \\
LongBenchV2-UserGuide & Technical manuals & Single-document long-document retrieval & 88 & 22 & Option-level binary verification & Binary accuracy \\
LongBenchV2-Gov & Government documents & Cross-document long-context QA & 21 & 44 & Multiple-choice & Option accuracy \\
\bottomrule
\end{tabular}
\caption{Evaluation datasets and units. UserGuide is reported at the level of 88 option-level binary verification instances over 22 technical manuals; Gov is evaluated as 21 multiple-choice questions over 44 government documents.}
\label{tab:appendix_benchmarks}
\end{table*}

\begin{table*}[!t]
\centering
\small
\setlength{\tabcolsep}{4pt}
\begin{tabular}{lrrr}
\toprule
Dataset & \# Docs. & Total tokens & Avg. tokens/doc. \\
\midrule
MDAQA & 792 & 8,256,933 & 10,425.42 \\
ScholarQA & 137 & 1,422,119 & 10,380.43 \\
LongBenchV2-UserGuide & 22 & 1,784,355 & 81,107.05 \\
LongBenchV2-Gov & 44 & 3,940,371 & 89,553.89 \\
\bottomrule
\end{tabular}
\caption{Corpus statistics for the evaluated data. Document counts for LongBenchV2-Gov refer to document instances.}
\label{tab:dataset_statistics}
\end{table*}

\subsection{Shared Document Preprocessing}
\label{app:shared-settings}

All systems receive the same source documents after a common paragraph-level segmentation step.
Documents are segmented before method-specific indexing.
Paragraphs shorter than 200 tokens are merged with adjacent paragraphs within the same section, while paragraphs longer than 400 tokens are split into smaller units.
We do not merge across section boundaries, so each resulting unit remains aligned with the source document structure.

\subsection{Evaluation Details}
\label{app:evaluation-details}

\paragraph{Metric scope.}
The primary answer-quality metric follows the answer format of each benchmark: RAGAS Answer Correctness for ScholarQA and MDAQA, option-level binary accuracy for UserGuide, and multiple-choice accuracy for Gov.
RAGAS Semantic Similarity and Faithfulness are reported only for open-ended scientific QA, while Sufficient Context is reported across all four benchmarks as a retrieval-side evidence-completeness signal.

\paragraph{Aggregation.}
Scores are macro-averaged over each benchmark's evaluation units: the original questions for ScholarQA, MDAQA, and Gov, and the 88 option-level verification instances for UserGuide.

\paragraph{Evaluator.}
LLM-judged metrics use \texttt{Claude-\allowbreak Sonnet-\allowbreak 4-\allowbreak 6} as the evaluator.
Semantic Similarity is embedding-based, using \texttt{all-mpnet-\allowbreak base-\allowbreak v2}, and is not treated as an LLM-judged metric.

\paragraph{Grounding context.}
For Faithfulness and Sufficient Context, grounding context includes only source text retrieved through \texttt{fetch\_content}.
Navigation-only artifacts, including outlines, parent scopes, neighbor candidates, routing snippets, graph profiles, and candidate-node metadata, are excluded unless they are materialized as source text through \texttt{fetch\_content}.

\subsection{Implementation Details}
\label{app:implementation-details}

\paragraph{LLM settings.}
We use \texttt{DeepSeek-v4-Flash} for all LLM-based components except evaluation, for which we use \texttt{Claude-\allowbreak Sonnet-\allowbreak 4-\allowbreak 6}.
All LLM calls use temperature 0.

\paragraph{Embedding and hardware.}
Retrieval embeddings and RAGAS Semantic Similarity are computed with \texttt{sentence-\allowbreak transformers/\allowbreak all-\allowbreak mpnet-\allowbreak base-\allowbreak v2}~\citep{reimers-gurevych-2019-sentence}.
Embedding computation and non-LLM preprocessing were run on a machine equipped with a single NVIDIA GeForce RTX 3060 GPU.

\paragraph{Comparison protocol.}
For fair comparison, all systems operate on the same preprocessed source chunks and use their respective indexing and retrieval pipelines.

\subsection{Additional Result Tables}
\label{app:additional-results}

\paragraph{Secondary diagnostics.}
\label{app:secondary-diagnostics}

Table~\ref{tab:appendix_secondary_diagnostics} reports Faithfulness, Token-F1, and BERTScore-F1 for the open-ended scientific benchmarks.
These diagnostics complement the primary answer-quality and evidence-completeness measures; the main interpretation uses Correctness and, where available, Semantic Similarity for answer quality and Sufficient Context for retrieval-side evidence completeness.

\begin{table*}[!tbp]
\centering
\scriptsize
\setlength{\tabcolsep}{2.0pt}
\begin{tabular}{llrrrrrrrrr}
\toprule
Dataset & Metric & Proposed & \multicolumn{3}{c}{Standard passage retrieval} & \multicolumn{3}{c}{Graph-based RAG} & \multicolumn{2}{c}{Agentic retrieval} \\
\cmidrule(lr){3-3}\cmidrule(lr){4-6}\cmidrule(lr){7-9}\cmidrule(lr){10-11}
 & & Ours & BM25 & TF-IDF & Dense & LinearRAG & HippoRAG & LightRAG & IRCoT & A-RAG \\
\midrule
ScholarQA & Faith. & \underline{0.803} & 0.674 & 0.690 & 0.687 & 0.772 & 0.751 & 0.508 & \textbf{0.823} & 0.533 \\
 & Tok-F1 & \underline{0.299} & 0.211 & 0.197 & 0.203 & 0.137 & 0.239 & \textbf{0.316} & 0.250 & 0.260 \\
 & BERT-F1 & \textbf{0.727} & 0.616 & 0.616 & 0.617 & 0.522 & 0.618 & \underline{0.710} & 0.638 & 0.675 \\
\midrule
MDAQA & Faith. & 0.730 & 0.693 & 0.632 & 0.720 & 0.754 & \underline{0.821} & 0.639 & \textbf{0.835} & 0.451 \\
 & Tok-F1 & 0.134 & \textbf{0.273} & \underline{0.267} & 0.264 & 0.255 & 0.244 & 0.178 & 0.247 & 0.094 \\
 & BERT-F1 & \textbf{0.677} & 0.667 & 0.654 & 0.654 & 0.610 & 0.628 & \underline{0.672} & 0.628 & 0.653 \\
\bottomrule
\end{tabular}
\caption{Secondary diagnostics for open-ended scientific QA. Faithfulness assesses whether generated answers are supported by retrieved source text, while Token-F1 and BERTScore-F1 measure answer overlap with the reference. Bold indicates the best value within each dataset and metric; underline indicates the second-best value.}
\label{tab:appendix_secondary_diagnostics}
\end{table*}

\paragraph{Model backbone generalization.}
\label{app:backbone-generalization}

Table~\ref{tab:backbone_generalization} reports a supplementary backbone evaluation on four benchmark rows: ScholarQA, MDAQA, LongBench-v2 UserGuide, and LongBench-v2 Gov.
DeepSeek-v4-Flash is the main-experiment backbone; the remaining five backbones are supplementary runs.
The table reports Correctness for all four benchmark rows and Semantic Similarity only for ScholarQA and MDAQA, consistent with the metric scope used in the main experiments.

\begin{table*}[!tbp]
\centering
\scriptsize
\setlength{\tabcolsep}{1.6pt}
\begin{tabular}{llrrrrrr}
\toprule
Backbone & Benchmark & Ours Correct. & A-RAG Correct. & $\Delta$ Correct. & Ours Sim. & A-RAG Sim. & $\Delta$ Sim. \\
\midrule
DeepSeek-v4-Flash (main) & LB-v2 UserGuide & 0.784 & 0.704 & +7.96 & -- & -- & -- \\
 & LB-v2 Gov & 0.619 & 0.524 & +9.52 & -- & -- & -- \\
 & MDAQA & 0.618 & 0.596 & +2.20 & 0.688 & 0.650 & +3.80 \\
 & ScholarQA & 0.738 & 0.671 & +6.70 & 0.797 & 0.743 & +5.40 \\
\midrule
DeepSeek-v4-Pro & LB-v2 UserGuide & 0.759 & 0.670 & +8.81 & -- & -- & -- \\
 & LB-v2 Gov & 0.524 & 0.476 & +4.76 & -- & -- & -- \\
 & MDAQA & 0.600 & 0.592 & +0.74 & 0.639 & 0.640 & -0.07 \\
 & ScholarQA & 0.706 & 0.690 & +1.65 & 0.779 & 0.759 & +2.01 \\
\midrule
Qwen3.5-Flash & LB-v2 UserGuide & 0.750 & 0.670 & +7.95 & -- & -- & -- \\
 & LB-v2 Gov & 0.381 & 0.333 & +4.77 & -- & -- & -- \\
 & MDAQA & 0.522 & 0.497 & +2.49 & 0.602 & 0.590 & +1.20 \\
 & ScholarQA & 0.579 & 0.567 & +1.21 & 0.684 & 0.678 & +0.51 \\
\midrule
Qwen3.5-Plus & LB-v2 UserGuide & 0.716 & 0.659 & +5.69 & -- & -- & -- \\
 & LB-v2 Gov & 0.381 & 0.476 & -9.52 & -- & -- & -- \\
 & MDAQA & 0.536 & 0.545 & -0.86 & 0.603 & 0.619 & -1.60 \\
 & ScholarQA & 0.614 & 0.613 & +0.12 & 0.710 & 0.720 & -0.94 \\
\midrule
GPT-5.4-mini & LB-v2 UserGuide & 0.818 & 0.793 & +2.51 & -- & -- & -- \\
 & LB-v2 Gov & 0.571 & 0.524 & +4.76 & -- & -- & -- \\
 & MDAQA & 0.598 & 0.589 & +0.87 & 0.670 & 0.664 & +0.54 \\
 & ScholarQA & 0.672 & 0.645 & +2.73 & 0.761 & 0.734 & +2.64 \\
\midrule
GPT-5.5 & LB-v2 UserGuide & 0.739 & 0.720 & +1.81 & -- & -- & -- \\
 & LB-v2 Gov & 0.429 & 0.476 & -4.76 & -- & -- & -- \\
 & MDAQA & 0.637 & 0.645 & -0.83 & 0.712 & 0.696 & +1.57 \\
 & ScholarQA & 0.683 & 0.714 & -3.11 & 0.771 & 0.778 & -0.70 \\
\bottomrule
\end{tabular}
\caption{Supplementary backbone evaluation on ScholarQA, MDAQA, LongBench-v2 UserGuide, and LongBench-v2 Gov. DeepSeek-v4-Flash is the main-experiment backbone; the other five backbones are supplementary runs. Correctness is reported for all four benchmark rows, while Semantic Similarity is reported only for ScholarQA and MDAQA. $\Delta$ is Ours minus A-RAG in percentage points, so positive values indicate that the proposed method is higher than A-RAG.}
\label{tab:backbone_generalization}
\end{table*}

\FloatBarrier

\begin{strip}
\centering
\tiny
\setlength{\tabcolsep}{2.2pt}
\renewcommand{\arraystretch}{0.92}
\resizebox{\textwidth}{!}{%
\begin{tabular}{@{}lrrrrrrrrrrrrrrr@{}}
\toprule
\multirow{2}{*}{Method}
& \multicolumn{5}{c}{LLM calls}
& \multicolumn{5}{c}{Input tokens}
& \multicolumn{5}{c}{Output tokens} \\
\cmidrule(lr){2-6}\cmidrule(lr){7-11}\cmidrule(l){12-16}
& Sch. & MDA & UG & Gov & Agg.
& Sch. & MDA & UG & Gov & Agg.
& Sch. & MDA & UG & Gov & Agg. \\
\midrule
BM25 & 33 & 30 & 88 & 21 & 172 & 68,716 & 31,675 & 224,211 & 72,733 & 397,335 & 23,805 & 27,230 & 53,062 & 39,347 & 143,444 \\
TF-IDF & 33 & 30 & 88 & 21 & 172 & 63,949 & 28,506 & 217,039 & 65,205 & 374,699 & 25,733 & 26,876 & 52,140 & 34,316 & 139,065 \\
Dense & 33 & 30 & 88 & 21 & 172 & 62,791 & 30,780 & 199,851 & 65,888 & 359,310 & 29,503 & 26,668 & 57,340 & 45,137 & 158,648 \\
LinearRAG & 33 & 30 & 88 & 21 & 172 & 60,278 & 31,160 & 209,430 & 53,542 & 354,410 & 15,813 & 15,520 & 41,818 & 41,251 & 114,402 \\
HippoRAG & 9,315 & 45,885 & 12,190 & 25,388 & 92,778 & 8,229,196 & 22,866,263 & 13,445,893 & 16,662,882 & 61,204,234 & 2,479,526 & 3,538,711 & 3,844,319 & 36,599,663 & 46,462,219 \\
LightRAG & 3,042 & 52,083 & 8,337 & 25,392 & 88,854 & 11,727,484 & 183,905,883 & 32,177,169 & 110,075,722 & 337,886,258 & 2,251,294 & 28,743,875 & 8,026,486 & 90,369,607 & 129,391,262 \\
IRCoT & 110 & 124 & 284 & 132 & 650 & 324,568 & 261,885 & 848,807 & 452,924 & 1,888,184 & 31,136 & 27,816 & 66,550 & 58,644 & 184,146 \\
A-RAG & 245 & 300 & 498 & 282 & 1,325 & 2,547,668 & 3,261,745 & 4,013,608 & 4,878,828 & 14,701,849 & 104,243 & 88,005 & 176,559 & 100,510 & 469,317 \\
\textbf{Ours} & 676 & 1,322 & 1,196 & 924 & 4,118 & 10,482,696 & 13,806,858 & 21,654,655 & 48,429,355 & 94,373,564 & 492,397 & 992,740 & 872,623 & 1,179,004 & 3,536,764 \\
\bottomrule
\end{tabular}
}
\captionof{table}{Full end-to-end LLM usage by benchmark and method. Calls and token counts include indexing, retrieval, and answer generation, and exclude judge-only evaluation calls. Within each metric group, columns denote ScholarQA (Sch.), MDAQA (MDA), LongBench-v2 UserGuide (UG), LongBench-v2 Gov (Gov), and Aggregate (Agg.).}
\label{tab:cost_results}
\end{strip}

\section{Method Implementation Details}
\label{app:method-details}

This appendix describes the implementation used to instantiate the document-structured graph and the agentic retrieval workflow.
The main text presents the method in terms of graph construction, graph actions, and fetched evidence.
Here, we specify the concrete construction pipeline, node-view allocation, lateral-link construction, runtime retrieval contract, and corpus-specific instantiations.

\subsection{Document Graph Construction}
\label{app:graph-construction-details}

\paragraph{Layout parsing and hierarchy recovery.}
We parse each source document with MinerU, which provides layout regions, title attributes, and reading-order information.
These signals define a document-native hierarchy whose root is the document, whose leaves are paragraph-level text units, and whose internal nodes correspond to recovered headings, sections, chapters, or other structural scopes.
The hierarchy preserves source organization without requiring a fully induced semantic knowledge graph, and all recovered structural nodes remain in the graph.
The adaptive scope rule from the main text decides which nodes receive richer routing representations; it does not remove small headings or local sections.
Local structural nodes remain available as context and participate in bottom-up profile construction, while broader sections and document nodes receive higher-cost global profiles.

\paragraph{Paragraph evidence units.}
Paragraphs are the basic evidence units because they preserve local discourse structure and remain small enough to fetch for answer generation.
Very short paragraphs are merged with adjacent paragraphs in the same section.
We do not merge across section boundaries, since cross-section merging can blur the recovered document structure.
Very long paragraphs are split to keep evidence units within a manageable range.
In the experiments, paragraph units target roughly 200--400 tokens.

\paragraph{Structural scopes and profile budget.}
Let $T_d$ denote the recovered hierarchy for document $d$.
Its leaves are paragraph evidence units, and its internal nodes are structural scopes.
We define
\[
S_d=\mathrm{Internal}(T_d),
\]
where $S_d$ contains all recovered structural nodes.
A subset of these nodes receives richer routing profiles:
\[
\begin{aligned}
R_d ={}& \{\mathrm{root}(T_d)\} \\
&{}\cup \{u\in S_d:
h(u)\ge \tau_h \ \mathrm{or}\ \ell(u)\ge \tau_\ell\},
\end{aligned}
\]
where $h(u)$ measures subtree height and $\ell(u)$ measures the token span covered by the scope.
Nodes in $S_d\setminus R_d$ remain in the graph with lightweight bottom-up views.
Nodes in $R_d$, such as document nodes and major sections, serve as broader routing scopes and receive richer global profiles.
This separates structural inclusion from representation cost: the graph preserves the recovered hierarchy, while expensive summarization is concentrated on scopes that can support higher-level routing.

\begin{table*}[t]
\centering
\small
\setlength{\tabcolsep}{4pt}
\begin{tabular}{lll}
\toprule
Node type & Included in graph & Representation budget \\
\midrule
Paragraph & Yes & Source text, embedding, entities, keywords, path \\
Small heading / local section & Yes & Lightweight bottom-up profile \\
Major section / chapter & Yes & Aggregated profile plus optional global profile \\
Document & Yes & Global profile for broad routing \\
Community & Optional & Topical neighborhood; not direct evidence \\
\bottomrule
\end{tabular}
\caption{
Structural inclusion and representation budget in the document graph.
All recovered structural nodes remain in the graph, while richer global profiles are assigned to broad routing scopes.
}
\label{tab:app-profile-budget}
\end{table*}

\paragraph{Bottom-up node views.}
Each graph node is exposed to retrieval through a level-conditioned node view.
Paragraph nodes keep source-close fields: paragraph text, embedding, structural path, named entities, and keywords.
Named entities are extracted at the chunk level with NER, and keywords are obtained from word frequency after stopword filtering.

Structural node views are constructed bottom-up from their children.
For embeddings, we use max pooling over child representations.
Other fields, including topic cues, keywords, and structural paths, are aggregated from child profiles.
For broad routing scopes, including document nodes and qualifying major sections, this bottom-up view is augmented with an LLM-generated global profile containing a summary, key concepts, and major subtopics.
This keeps most graph nodes source-close and lightweight while giving broad scopes enough semantic detail for routing.

\begin{table}[H]
\centering
\scriptsize
\setlength{\tabcolsep}{2pt}
\renewcommand{\arraystretch}{1.04}
\begin{tabularx}{\columnwidth}{@{}>{\RaggedRight\arraybackslash}p{0.19\columnwidth}>{\RaggedRight\arraybackslash}p{0.34\columnwidth}>{\RaggedRight\arraybackslash}X@{}}
\toprule
View field & Source & Use in retrieval \\
\midrule
Text & Paragraph content & Fetched evidence and local matching \\
Embedding & Paragraph or pooled child embeddings & Semantic retrieval and similarity links \\
Structural path & Recovered hierarchy & Scope-aware navigation \\
Entities & NER over paragraph text & Entity-based lateral links \\
Keywords & Frequency extraction with stopword filtering & Lexical routing cues \\
Summary & LLM profile for broad scopes & High-level routing \\
Key concepts & LLM profile for broad scopes & Topic matching and routing \\
Major subtopics & LLM profile for broad scopes & Outline-like scope selection \\
\bottomrule
\end{tabularx}
\caption{
Example node-view fields used in the document graph.
Paragraph nodes remain source-close, smaller structural nodes use bottom-up profiles, and broad routing scopes receive higher-cost global profiles.
}
\label{tab:app-node-view-fields}
\end{table}

\paragraph{Lateral links and topical communities.}
In addition to hierarchical containment edges, we add lightweight lateral links when the corresponding signals are available.
Local adjacency links connect neighboring paragraph or section nodes that share the same parent scope, using recovered reading order and sibling structure.
Citation links are extracted from chunk-level citation mentions and attached to both the local text region and relevant ancestor scopes, allowing the controller to recover citation-related context at multiple granularities.
Entity links are constructed from a chunk--entity matrix, so chunks that share salient entities can be considered potential evidence neighbors.

Semantic similarity links are mainly constructed over higher-level node profiles, such as document profiles or major-section profiles, where similarity is more useful for routing across broad regions.
For broad-scope association, we build topical communities with Leiden community detection.
In scientific paper collections, communities are typically built over document-level profiles because documents are the main cross-source units.
In long-document corpora, communities are typically built over major-section profiles because major sections are more useful cross-region units than whole documents.
Community structure may appear as optional community nodes or as topical neighborhood relations.
In both cases, communities support routing rather than final answer grounding.

\begin{table}[H]
\centering
\scriptsize
\setlength{\tabcolsep}{2pt}
\renewcommand{\arraystretch}{1.04}
\begin{tabularx}{\columnwidth}{@{}>{\RaggedRight\arraybackslash}p{0.28\columnwidth}>{\RaggedRight\arraybackslash}p{0.36\columnwidth}>{\RaggedRight\arraybackslash}X@{}}
\toprule
Link type & Construction signal & Typical level \\
\midrule
Hierarchical containment & Recovered document hierarchy & paragraph--section--document \\
Local adjacency & Reading order and sibling structure & paragraph / section \\
Citation & Citation mentions and ancestor propagation & chunk, section, document \\
Entity & Chunk--entity matrix & chunk \\
Semantic similarity & Node profile similarity & major section / document \\
Topical community & Leiden over broad-scope profiles & document or major section \\
\bottomrule
\end{tabularx}
\caption{
Structural and lateral links used by the document graph.
Available link types depend on the corpus; for example, long user-guide documents do not contain citation links.
}
\label{tab:app-graph-links}
\end{table}

\FloatBarrier
\subsection{Agentic Graph Retrieval Workflow}
\label{app:agentic-workflow}

\begin{figure*}[!t]
    \centering
    \includegraphics[width=\textwidth]{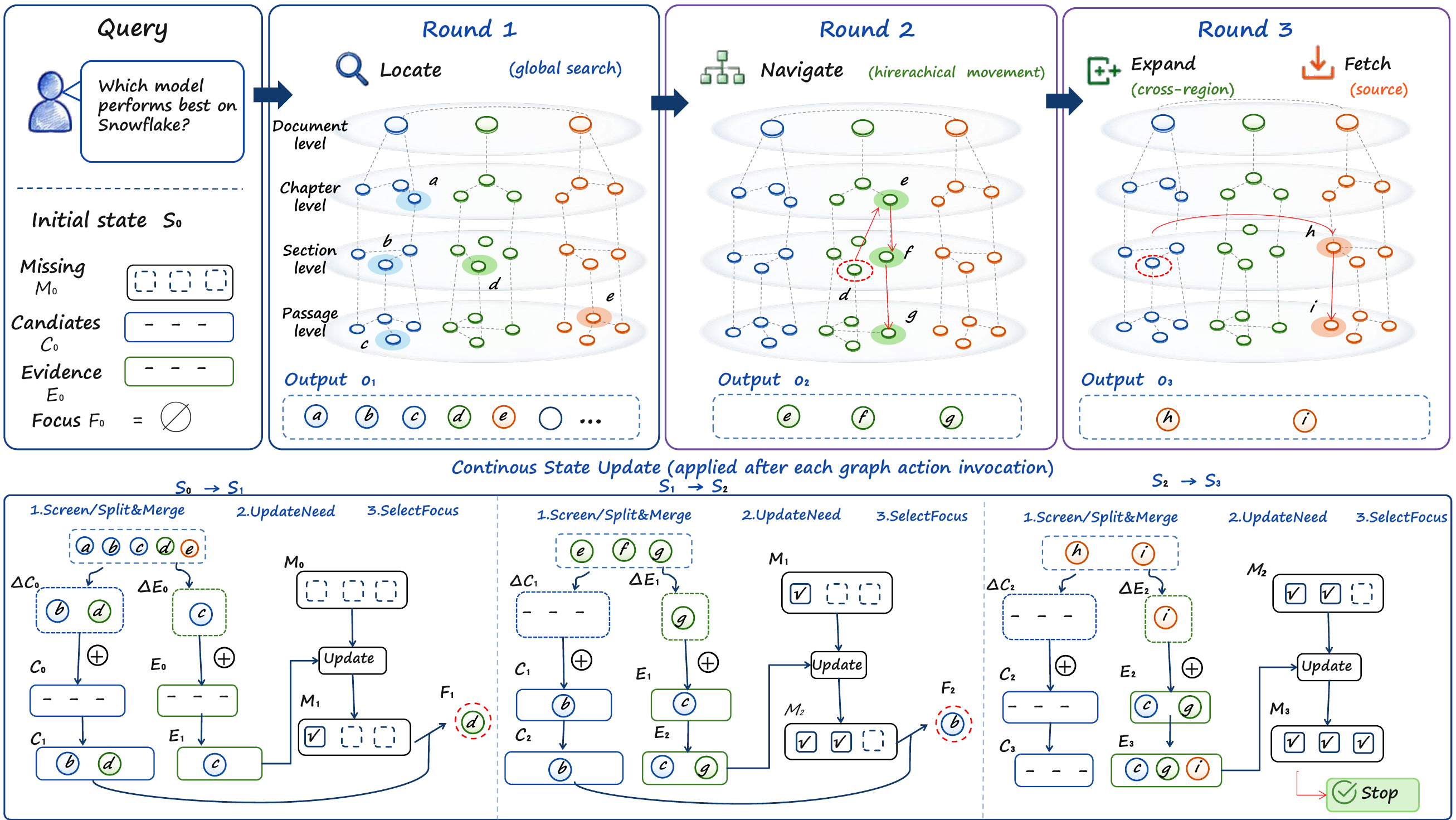}
    \caption{Detailed query-time retrieval workflow in \system. The controller maintains a retrieval note, selects a graph action from the constrained action interface, executes the corresponding backend primitive, screens returned graph observations, revises unresolved evidence needs and the next focus, and admits only fetched source text as answer-grounding evidence.}
    \label{fig:app-query-overview}
\end{figure*}

Figure~\ref{fig:app-query-overview} expands the query-time retrieval loop summarized in the main framework overview.

\paragraph{Controller implementation.}
The query-time controller is implemented as an LLM skill prompt that follows a fixed graph-retrieval workflow.
In the main experiments, the controller LLM is DeepSeek-v4-Flash.
In backbone experiments, the controller LLM changes with the corresponding backbone setting.
The controller does not call arbitrary tools; it selects from a fixed set of graph actions and updates a retrieval note after each step.

\paragraph{Retrieval note.}
The retrieval note is the controller's loop state.
It records the query, answer goal, discovered information, unresolved needs, next intended graph move, and answer readiness.
The note is stored as compact natural-language state:

\begin{verbatim}
{
  "query": "...",
  "rough_goal": "...",
  "found_so_far": "...",
  "missing_so_far": "...",
  "next_step": "...",
  "can_answer": false
}
\end{verbatim}

The implementation realizes the main-text state $s_t=(F_t,C_t,E_t,M_t)$ through this note: \texttt{next\_step} stores the next focus or graph move, \texttt{found\_so\_far} stores candidate scopes and fetched-evidence references, and \texttt{missing\_so\_far} stores unresolved evidence needs; \texttt{can\_answer} records the stopping readiness.
The field \texttt{found\_so\_far} summarizes discovered scopes, candidate nodes, and fetched evidence references.
The field \texttt{missing\_so\_far} records unresolved evidence needs, such as a missing definition, comparison target, broader context, or supporting source.
The field \texttt{can\_answer} is set to true only when the fetched evidence is sufficient for answer generation.

\paragraph{Graph action interface.}
\label{app:graph-action-interface}
At each loop iteration, the controller selects one of the four abstract actions defined in the main text.
Given the current focus, retained candidates, target granularity, and corpus-supported relations, the execution layer realizes that action with one backend primitive or a short sequence of primitives.
The primitives operate on stable node identifiers and expose search, structural traversal, outline inspection, lateral expansion, and source materialization separately.
Their outputs are parsed into paragraph or structural node observations before relevance screening and state update; thus, the abstract actions specify retrieval intent rather than a one-to-one mapping to code-level tools.
\textsc{Answer} remains a terminal generation step outside the graph-action space.

\begin{table*}[t]
\centering
\scriptsize
\setlength{\tabcolsep}{2.5pt}
\renewcommand{\arraystretch}{1.08}
\begin{tabularx}{\textwidth}{
    @{}
    >{\RaggedRight\arraybackslash}p{0.09\textwidth}
    >{\RaggedRight\arraybackslash}p{0.14\textwidth}
    >{\RaggedRight\arraybackslash}p{0.35\textwidth}
    >{\RaggedRight\arraybackslash}X
    @{}
}
\toprule
\textbf{Action} & \textbf{Primitive} & \textbf{Execution} & \textbf{Output and downstream use} \\
\midrule
\textsc{Locate}
& \texttt{global\_search}
& Searches the selected graph level with the current evidence need, without a structural scope filter.
& Ranked node cards with node id, level, score, title or snippet, and available scope metadata; these initialize or refresh the candidate set. \\
\addlinespace[1pt]
\multirow{4}{*}{\textsc{Navigate}}
& \texttt{scoped\_search}
& Applies document, community, or section-scope filters to search descendants of the active scope; candidate-node filters instead rerank a bounded set from earlier steps.
& Ranked nodes restricted to the supplied scope or candidate set, used to select a narrower focus. \\
& \texttt{expand\_up}
& Resolves supplied node ids to requested parent or ancestor levels through deterministic hierarchy links.
& Parent structural-node cards grouped by seed and target level, used to broaden the current focus. \\
& \texttt{expand\_down}
& Resolves a structural focus to child or descendant nodes at requested lower levels.
& Descendant paragraph or structural-node cards, used for direct downward navigation. \\
& \texttt{inspect\_outline}
& Reads the local hierarchy below a selected node, exposing child titles, nesting, and optional paragraph entries; the controller chooses an entry for subsequent resolution.
& Outline entries used to identify the next node or scope; the entries themselves remain navigation context. \\
\addlinespace[1pt]
\textsc{Expand}
& \texttt{expand\_neighbors}
& Traverses the requested lateral or local edge types from one or more seed nodes.
& Neighbor-node cards together with edge type, weight, and relation metadata, which are screened as new candidates. \\
\addlinespace[1pt]
\textsc{Fetch}
& \texttt{fetch\_content}
& Resolves selected node ids to source content: paragraph nodes yield their full text, whereas structural nodes yield routing fields and associated representative paragraphs.
& Source-bearing paragraph nodes can enter evidence after screening; structural routing fields remain navigation context. \\
\bottomrule
\end{tabularx}
\caption{
Execution contract of the backend primitives underlying the four abstract graph actions.
An abstract action may invoke one primitive or compose a short sequence before returning normalized graph nodes.
Global and scoped search share \texttt{search\_nodes.py}; upward and downward traversal share \texttt{expand\_structural.py}; the remaining primitives use \texttt{get\_outline.py}, \texttt{expand\_neighbors.py}, and \texttt{fetch\_content.py}.
}
\label{tab:app-action-contract}
\end{table*}

Evidence admission depends on the returned node content rather than on the action name.
Search and expansion outputs share stable node identifiers and levels, allowing a returned candidate to be passed directly to a later structural, lateral, or content operation.
After relevance screening, a paragraph node that carries original source text may enter $\Delta E_t$; structural nodes, outline entries, routing profiles, and snippet-only node cards remain navigation context or candidates.
In the MultiGov backend, search, outline, and neighbor primitives return such candidate records, so full paragraph text is ordinarily materialized through \texttt{fetch\_content}; backends that already return source-bearing paragraph nodes can admit them without an additional fetch.

\paragraph{Retrieval loop and run records.}
For each query, the controller initializes a retrieval note and enters an iterative loop.
At each iteration, it reads the current note, selects the most important evidence need, chooses one graph action, executes that action, and updates the note.
Every run is recorded under a \texttt{.retrieval\_run/} directory.
Each graph action creates an immutable step artifact under \texttt{step\_artifacts/}.
If \textsc{FetchContent} retrieves source text that may support the answer, the run also creates one or more evidence artifacts under \texttt{evidence\_artifacts/}.

A step artifact records the action, why it was selected, the relevant tool references, what happened, new findings, the updated retrieval note, and any evidence created by the step.
This record makes the retrieval trajectory auditable: later steps can be traced back to the graph action and state update that produced them.

\paragraph{Evidence artifacts and grounding boundary.}
The workflow separates navigation signals from answer-grounding evidence.
Search results, outlines, parent scopes, and neighbor nodes can guide later retrieval, but they are not used directly as final-answer evidence.
A node becomes answer-grounding evidence only after \textsc{FetchContent} materializes its source text and writes an evidence artifact.

Each evidence artifact records an evidence identifier, source step, node id and level, fetched text, usefulness rationale, and a reference to the underlying retrieval output.
The final answer is written in the same skill context as retrieval, but its grounding is restricted to fetched evidence artifacts.
This prevents the generator from relying on unfetched snippets or merely structural associations.

\paragraph{Stopping condition.}
The retrieval loop exits when the retrieval note indicates that the answer can be written from existing evidence artifacts.
If the fetched evidence does not yet support the answer, the controller continues searching, expanding, inspecting, or fetching according to the unresolved needs recorded in the note.
The implementation can also stop when no useful new graph move is available or when a run-specific retrieval limit is reached.

\FloatBarrier
\subsection{Corpus-Specific Instantiations}
\label{app:corpus-instantiations}

The same graph and retrieval abstraction is instantiated differently across corpus types.
Scientific paper collections have many documents and citation signals, so document-level profiles and document-level topical communities are useful for broad routing.
Long structured documents contain fewer documents but many major sections, so major-section profiles and section-level communities are more useful for cross-region navigation.

\begin{table}[H]
\centering
\scriptsize
\setlength{\tabcolsep}{2pt}
\renewcommand{\arraystretch}{1.04}
\begin{tabularx}{\columnwidth}{@{}>{\RaggedRight\arraybackslash}p{0.28\columnwidth}>{\RaggedRight\arraybackslash}p{0.22\columnwidth}>{\RaggedRight\arraybackslash}X@{}}
\toprule
Corpus type & Bottom evidence unit & Common structural and lateral signals \\
\midrule
Scientific paper collections & paragraph / chunk & section, document, citation, entity, similarity, community \\
Long structured documents & paragraph / chunk & section, major section, document, entity, similarity, local adjacency, community \\
\bottomrule
\end{tabularx}
\caption{
Corpus-specific instantiations of the document graph.
The action space is shared across settings, but action modes and available lateral links depend on the corpus.
}
\label{tab:app-corpus-instantiation}
\end{table}

\section{Ablation Implementation Details}
\label{app:ablation-details}

\begin{table}[H]
\centering
\scriptsize
\setlength{\tabcolsep}{1.6pt}
\renewcommand{\arraystretch}{1.04}
\resizebox{\columnwidth}{!}{%
\begin{tabular}{@{}lccccc@{}}
\toprule
Variant & Multi-level & Structural & Lateral & State & Fetch \\
\midrule
Full & \checkmark & \checkmark & \checkmark & \checkmark & \checkmark \\
w/o Hierarchy Search & -- & -- & -- & \checkmark & \checkmark \\
w/o Vertical & \checkmark & -- & \checkmark & \checkmark & \checkmark \\
w/o Lateral & \checkmark & \checkmark & -- & \checkmark & \checkmark \\
w/o State Control & \checkmark & \checkmark & \checkmark & -- & \checkmark \\
w/o Evidence Fetch & \checkmark & \checkmark & \checkmark & \checkmark & -- \\
\bottomrule
\end{tabular}
}
\caption{Ablation variants and removed capabilities. Columns denote multi-level search, structural scope/vertical navigation, lateral graph expansion, dynamic retrieval state, and the fetched-evidence boundary.}
\label{tab:appendix_ablation_variants}
\end{table}

All ablations are independent workflow variants that keep the datasets, LLM, evaluation scripts, retrieval backend, index, embeddings, graph construction, and evaluation pipeline fixed.
They change only the controller-visible retrieval actions or evidence-control rules, and they do not add sample filtering or alter the graph, index, embedding cache, or backend retrieval store.

\paragraph{Full setting.}
The full workflow maintains a retrieval note with the query goal, discovered scopes and candidates, fetched evidence, missing evidence needs, next action, and answer readiness.
Each step selects one graph action, writes a step artifact, and admits final-answer evidence only after \texttt{fetch\_content} creates an evidence artifact.
Thus the full setting exposes multi-level search, scoped search, upward expansion, outline inspection, lateral neighbor expansion, and content fetching.

\paragraph{Action ablations.}
w/o Hierarchy Search exposes only chunk search, optional candidate reranking, and chunk fetching.
w/o Vertical keeps multi-level search, reranking, lateral expansion, and fetching, but removes scope filters, upward expansion, and outline inspection.
w/o Lateral keeps hierarchical search, structural navigation, outline inspection, and fetching, but removes neighbor, diffusion, citation, similarity, and same-document adjacency expansion.

\paragraph{State and evidence-control ablations.}
w/o State Control removes retrieval-note-driven action selection and stopping, replacing it with a fixed sequence of search, structural expansion, lateral expansion, candidate construction, fetching, and answer generation.
w/o Evidence Fetch keeps the dynamic loop and step artifacts but allows search, outline, structural-expansion, and neighbor-expansion records to enter the answer context without \texttt{fetch\_content}.

For ScholarQA and MDAQA, graph levels are community, document, section, and chunk; for LongBenchV2-UserGuide, they are document, chapter or section, and local evidence scopes.
The main-text ablation figure reports Correctness drops, while Table~\ref{tab:appendix_ablation_per_dataset} reports per-dataset Correctness and Semantic Similarity drops where each metric is defined.
Semantic Similarity is omitted for LongBenchV2-Gov because the benchmark uses multiple-choice outputs.

\begin{table}[H]
\centering
\scriptsize
\setlength{\tabcolsep}{1.5pt}
\begin{tabular}{@{}lcccc@{}}
\toprule
Variant & ScholarQA & MDAQA & \shortstack{LB-v2\\UserGuide} & \shortstack{LB-v2\\Gov} \\
& C/S & C/S & C/S & C/S \\
\midrule
w/o Hierarchy Search & 4.76/2.70 & 4.87/5.14 & 0.94/0.18 & 9.52/-- \\
w/o Vertical & 0.97/0.76 & 1.70/1.33 & 1.14/1.12 & 9.52/-- \\
w/o Lateral & 2.24/1.34 & 1.50/3.66 & 1.14/1.12 & 23.81/-- \\
w/o State Control & 3.18/3.03 & 2.39/4.63 & 3.41/3.39 & 23.81/-- \\
w/o Evidence Fetch & 2.82/2.22 & 2.70/3.86 & 4.55/4.53 & 28.57/-- \\
\bottomrule
\end{tabular}
\caption{Per-dataset ablation drops in percentage points (Correctness/Similarity). Similarity is not reported for LongBenchV2-Gov.}
\label{tab:appendix_ablation_per_dataset}
\end{table}

\end{document}